\documentclass{article}

\usepackage{arxiv}
\usepackage{svg}
\usepackage[utf8]{inputenc} 
\usepackage[T1]{fontenc}    
\usepackage{hyperref}       
\usepackage{url}            
\usepackage{booktabs}       
\usepackage{amsfonts}       
\usepackage{nicefrac}       
\usepackage{microtype}      
\usepackage{lipsum}
\usepackage{graphicx}
\graphicspath{ {./images/} }
\usepackage{amsmath}
\usepackage{algorithm}
\usepackage{algpseudocode}
\usepackage{adjustbox}
\usepackage{color}
\usepackage{graphicx,subfigure}
\usepackage{comment}
\usepackage{bm}

\usepackage{makecell}
\usepackage[utf8]{inputenc} 
\usepackage[T1]{fontenc}    
\usepackage{hyperref}       
\usepackage{url}            
\usepackage{booktabs}       
\usepackage{amsfonts}       
\usepackage{nicefrac}       
\usepackage{microtype}      
\usepackage{lipsum}
\usepackage{graphicx}
\graphicspath{{media/}}     
\usepackage{geometry}
\graphicspath{ {./images/} }

\title{AE-ViT: Stable Long-Horizon Parametric Partial Differential Equations Modeling}

\author{
 Iva Mikuš \\
  University of Zagreb Faculty of Electrical Engineering and Computer Science\\
  \texttt{ } \\
   \And
 Boris Muha \\
  University of Zagreb Faculty of Science Department of Mathematics\\
  \texttt{ } \\
  \And
 Domagoj Vlah \\
  University of Zagreb Faculty of Electrical Engineering and Computer Science\\
  \texttt{ } \\
}

\begin{document}
\maketitle
\begin{abstract}
Deep Learning Reduced Order Models (ROMs) are becoming increasingly popular as surrogate models for parametric partial differential equations (PDEs) due to their ability to handle high-dimensional data, approximate highly nonlinear mappings, and utilize GPUs. Existing approaches typically learn evolution either on the full solution field, which requires capturing long-range spatial interactions at high computational cost, or on compressed latent representations obtained from autoencoders, which reduces the cost but often yields latent vectors that are difficult to evolve, since they primarily encode spatial information.
Moreover, in parametric PDEs, the initial condition alone is not sufficient to determine the trajectory, and most current approaches are not evaluated on jointly predicting multiple solution components with differing magnitudes and parameter sensitivities. To address these challenges, we propose a joint model consisting of a convolutional encoder, a transformer operating on latent representations, and a decoder for reconstruction. The main novelties are joint training with multi-stage parameter injection and coordinate channel injection.
Parameters are injected at multiple stages to improve conditioning. Physical coordinates are encoded to provide spatial information. This allows the model to dynamically adapt its computations to the specific PDE parameters governing each system, rather than learning a single fixed response. Experiments on the Advection-Diffusion-Reaction equation and Navier-Stokes flow around the cylinder wake demonstrate that our approach combines the efficiency of latent evolution with the fidelity of full-field models, outperforming DL-ROMs, latent transformers, and plain ViTs in multi-field prediction, reducing the relative rollout error by approximately $5$ times.
\end{abstract}


\section{Introduction}
Speeding up calculations of the solution to parametric time-dependent PDEs is important in many applications, such as hemodynamics \cite{Buoso2019}, \cite{YE2024112639} and aerodynamics \cite{Dowell1999ReducedOM}. 
Precisely, for parametric PDE
\begin{align*}
    \begin{cases}
    \phi_t + \mathcal{L}(\phi; \lambda) = f(\cdot;\lambda) \text{ in $\Omega(\lambda)$ } \\
    \mathcal{B}( \phi ;\lambda)  = 0 \text{ on $\partial \Omega(\lambda)$} \\
    \phi(0,\cdot) = \phi_0(\cdot;\lambda),
    \end{cases}
\end{align*}

where $\mathcal{L}$ and $\mathcal{B}$ are parameter-dependent differential and boundary operators, $\lambda$ denotes the parameters, and $\phi_0$ and $f$ are the initial data and volume force, respectively, which may depend on parameters. Moreover, the domain also may depend on parameters, which adds additional geometric nonlinearity to the problem and is motivated by our long-term aim to adapt these methods to fluid-structure interaction problems. The goal is to design a surrogate model that quickly and effectively maps parameters to solutions, namely, to design an approximation of the mapping
\begin{align}
    (\lambda, t ) \rightarrow \phi(\cdot, t; \lambda).
\end{align}
Due to the success of deep learning in various domains and the universal approximation capabilities of neural networks, there is increasing interest in utilizing deep learning to obtain parametric PDE surrogates of evolutionary PDE.

In terms of deep learning, the solution for the parameters $\lambda$ at time $t$ is usually interpreted as an image, graph, or point cloud.  In this work, we focus on solutions that are either naturally on a rectangular grid or are interpolated to a rectangular grid so that convolutional neural networks can be used. Another assumption is that the data are collected with a fixed time step $\Delta t$. 

Many surrogate and operator-learning models for evolutionary PDEs, including DeepONet \cite{osti2281727}, \cite{He2024}, and Fourier Neural Operators (FNO) \cite{li2021fourierneuraloperatorparametric}, learn mappings from input functions (e.g., initial conditions or forcing terms) to solution trajectories. In these frameworks, physical or material parameters such as density, viscosity, or domain properties are typically incorporated only as part of the input function, for example, as constant fields or concatenated vectors, rather than as separate explicitly conditioned features. In cases of parametric PDE where parameters are material, fluid, domain properties, etc., such approaches may underperform because they do not have enough information about parameters. Furthermore, the initial condition may be the same for all parameter instances, so without parameter information, a neural network cannot differentiate the trajectories. In this work, we tackle this challenge by employing a fully convolutional neural encoder and decoder coupled with a vision transformer (ViT) \cite{dosovitskiy2021imageworth16x16words}, which we call AE-ViT. The encoder, the decoder, and ViT are enriched with parameter information through parameter injections across the model. For finer spatial awareness, we investigate the effect of coordinate positional channels on AE-ViT performance. To enhance autoregressive
stability, we use short scheduled sampling \cite{bengio2015scheduledsamplingsequenceprediction}.  Additionally, we demonstrate that our model is capable of learning multiple components of the solution jointly and that autoregressive relative errors remain stable beyond the scheduled sampling training window. 

We focus on the most dominant setting of in-distribution rollouts from unseen values of parameters within the training parameter range. While the extrapolation across parameters remains challenging, our goal here is stable long-horizon simulation within the calibrated regime. 

\subsection{Related work} 
In this subsection, we will categorize related deep learning methods for evolutionary parametric PDE into autoencoder-based approaches and models trained only on full-field solutions. 
\subsubsection{Autoencoder-based approaches}
These works use an autoencoder to first obtain the latent representation of the solution. Usually, one constructs the encoder to obtain the latent representation, the processor to obtain the predicted latent evolution, and the decoder for decoding the predicted latent representation.   
In \cite{NIKOLOPOULOS2022104652}, \cite{Franco_2022} a multi-layer perceptron (MLP) is used to map parameters (which may include time) to the latent representation.
A Transformer architecture  \cite{NIPS20173f5ee243} can be used to learn latent evolution, see \cite{Solera-Rico2024-pn}, \cite{hemmasian2023reduced}. The parameters are not encoded in the architecture but are inferred implicitly from the available trajectory. 
Another approach is to model latent space evolution under the law $l'(t; \lambda) = f(t, l(t); \lambda)$, where $f$ is approximated by the neural network and the evolution of the latent space is obtained by classical ODE solvers \cite{fresca2021comprehensive}. Fourier Neural Operator (FNO) \cite{li2021fourierneuraloperatorparametric} is used to model latent evolution \cite{LI2025113705}. This model has not yet been adapted to the parametric setting.
 
\subsubsection{Evolution on full-field}
 FactFormer \cite{li2023scalabletransformerpdesurrogate} is a transformer for PDE surrogate modeling that uses factorized axial attention. Instead of computing full attention across all grid points (which is unstable/expensive for high-res PDEs), they break it down into 1D factorized kernel integrals along each axis. This leverages the low-rank structure of PDE operators and reduces complexity. Another approach is to use vectorized conditional neural fields \cite{xie2022neuralfieldsvisualcomputing}, combined with a transformer to predict the whole time trajectory in one step \cite{hagnberger2024vectorizedconditionalneuralfields}.  

In contrast to scalable-attention approaches such as FactFormer and continuous-time neural fields such as VCNeF \cite{hagnberger2024vectorizedconditionalneuralfields}, our method emphasizes joint, parameter-aware operator learning. We show that this design is particularly effective for multi-field, scale-imbalanced parametric PDEs, where existing methods either separate training objectives or overlook the challenges of parameter conditioning.
\subsection{Positioning Our Work and Main Novelties}
Established operator-learning methods such as DeepONet and FNO are typically applied to initial-condition–to-trajectory settings rather than parameterized PDEs. Most existing parametric reduced-order models (ROMs) and latent evolution approaches (e.g., DL-ROM, Neural ODEs, and latent transformers) train encoders and evolution models separately. Our approach achieves improved predictive performance compared to these methods. We hypothesize that separate training can result in latent spaces that prioritize reconstruction accuracy over predictive robustness, which limits generalization across parameter spaces. In contrast, our approach emphasizes joint, parameter-aware operator learning, directly addressing this limitation. We combine the strengths of multiple paradigms: a convolutional encoder-decoder to reduce spatial resolution and a Vision Transformer (ViT) to capture non-local interactions. This design requires fewer trainable parameters than purely transformer- or autoencoder-based methods while maintaining stability over long-horizon autoregressive rollouts. Additionally, most latent evolution models are trained on a one-dimensional latent vector.

Models on full-field, which rely only on convolution for processing spatial information, can fail to capture non-local correlation due to the inherently limited receptive field of convolutions. In order to mitigate this, transformer-based architectures such as FactFormer \cite{li2023scalabletransformerpdesurrogate} and VCNeF have pushed scalability and temporal flexibility, respectively. 
VCNeF uses the initial condition and PDE parameters as inputs, enabling spatial and temporal interpolation as well as zero-shot super-resolution. By querying solutions directly as a function of time and space, VCNeF avoids autoregressive rollout and instead fits a global representation of the solution within a fixed temporal window. Long-horizon behavior under compounding error is not explicitly assessed.
While this formulation provides flexibility in spatial resolution and efficient interpolation, existing evaluations of VCNeF focus on relatively short temporal horizons, typically limited to a few dozen solver time steps. FactFormer leverages factorized axial attention to scale to large grids but does not incorporate parameter conditioning and remains focused on initial condition trajectory prediction. VCNeF conditions on initial conditions and parameters while enabling continuous-time prediction and temporal super-resolution, but is memory-heavy since it uses an attention mechanism \cite{NIPS20173f5ee243} on many query points over space and time. By using a convolutional encoder to reduce resolution and ViT as a processor, we model non-local interactions with much less memory and computational resources.

The main contributions of this work can be summarized as the following:
\begin{itemize}
    \item Joint training of encoder, processor, and decoder with multi-stage parameter injection
    \item Injection of coordinate channels to obtain better spatial awareness
    \item Multi-field training to obtain the solution of the system of PDEs simultaneously 
    \item Accurate long-term autoregressive rollout predictions despite a short training scheduled sampling window of length $4$
    \item Theoretical motivation and intuitive interpretation of the model through a kernel regression perspective
\end{itemize}

\section{Method}
\begin{figure}
    \centering
    \includegraphics[width=0.35\linewidth]{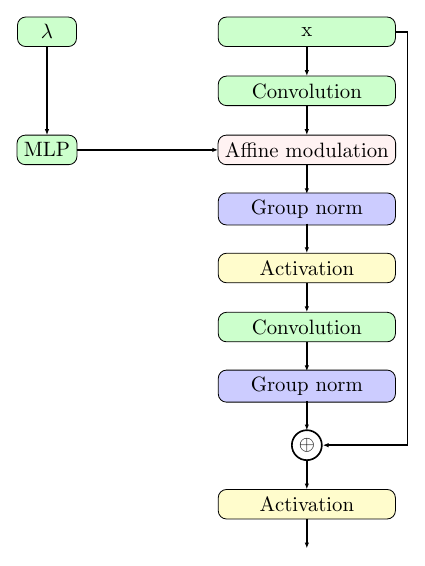}
    \caption{ResNet block with parameter injection. Parameters are passed through MLP and injected into the ResNet block through affine modulation after the first convolution in the residual block. }
    \label{fig:resnet_block}
\end{figure}

\begin{figure}
    \centering
    \includegraphics[width=0.7\linewidth]{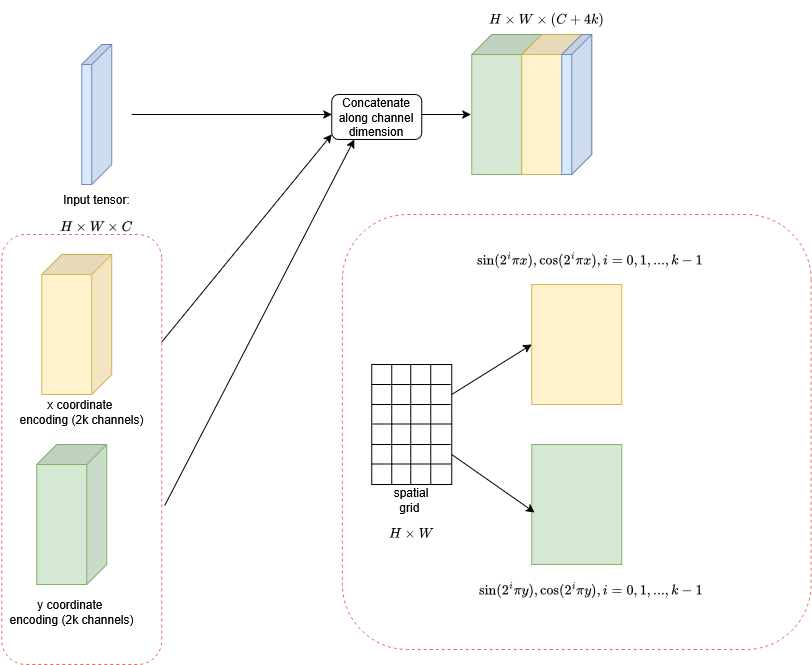}
    \caption{Coordinate encoding. Spatial Fourier features are constructed by applying sinusoidal functions at multiple frequencies to the coordinate grid $(x,y)$. The resulting $2k$ channels per axis are concatenated with the original feature map, yielding an augmented tensor of size $H \times W \times (C+4k)$, where $H$  and $W$, denote grid height and width and $C$ is the number of solution components. }
    \label{fig:coordinateEnc}
\end{figure}

The training set consists of parameter-solution pairs $(\lambda^{(i)}, \phi_j^{(i)}), i = 1 \ldots N_S, j = 1 \ldots N_T$, where $N_S$ is the number of training simulations, and $N_T$ is the number of training steps per simulation. We assume all simulations are sampled with the same time step $\Delta t$, $t_j = j \Delta t$ and $\phi_j^{(i)} = \phi(:, t_j; \lambda^{(i)})$ is the solution for parameter $\lambda^{(i)}$ of simulation $i$ and time $t_j$.
The goal is to construct a neural network that will for given parameters $\lambda$ calculate solution $\phi(:,t_j , \lambda)$ for $j=1,\ldots N_T$. Our proposed AE-ViT is a neural network that consists of the convolutional encoder,  the vision transformer and the decoder. Parameter-injection modules generally do not include time as a parameter. While incorporating time can aid short-term training, it tends to degrade performance in time-extrapolation regimes, which is why our design omits it.
\paragraph{Scheduled sampling.}
Since the model is doing autoregressive rollout, it is important to be able to predict from its own outputs, not just from correct data. In order to do so, we use scheduled sampling with fixed window. More precisely, during one training step we consider a window of consecutive time steps. For each sample in the window, with probability $p$,  correct $\phi^t$ is fed into the model instead of the  model prediction $\hat{\phi}_t$. This mechanism exposes the model to its own prediction errors during training, improving robustness and reducing error accumulation at inference time. In order to stabilize training, $p$ is large at the beginning of the training and it decreases as training progresses, feeding more and more model's own predictions as inputs. Scheduled sampling does not increase GPU usage per training step, but it increases training time. In this work, we use inverse sigmoid decay, defined as 
\begin{align*}
     p_i = \frac{k}{k + \exp(i/k)},
\end{align*}
where $i$ is optimization step, and $k = 0.3T_{max}/\log(0.3T_{max})$, with $T_{max}$ total number of training steps. 
\paragraph{Encoder structure.} In order to reduce spatial resolution and obtain a latent representation, a fully convolutional encoder is constructed. In order for the encoder to be aware of the PDE parameters feature-wise linear affine modulation (FiLM) transformation is used \cite{perez2017filmvisualreasoninggeneral}, \cite{farenga2024latentdynamicslearningnonlinear}, such that each hidden state $h$ is transformed as 
\begin{align*}
    h \xleftarrow{} \alpha(\lambda) \odot h \oplus \beta(\lambda),
\end{align*}
where $\oplus$ and $\odot$ are elementwise addition and multiplication, $\alpha, \beta $ are fully-connected MLPs mapping parameters $\lambda$ to a vector in $\mathbb R^{n_c}$ , and $n_c$ is the number of channels in the layer. Each channel $c_i$ is affinely modulated by parameter-dependent factor $\alpha(\lambda)_i$ and bias $\beta(\lambda)_i$. Additionally, we use ResNet blocks \cite{he2015deepresiduallearningimage}. Modified residual block is shown in the Figure \ref{fig:resnet_block}. We use Group Normalization \cite{wu2018groupnormalization} since it is batch size independent and stabilizes training of deep networks. In this setting, FiLM acts as a channel-wise reweighting mechanism rather than a pure affine transform.  The number of groups for a layer is set to be $\min(32, n_c/4)$. Such an encoder produces a tensor as its latent representation, thus preserving spatial relationships. The fully convolutional structure also reduces the number of training parameters.
\paragraph{Coordinate Encoding.}
In order for the model to have more spatial awareness, coordinate encoding channels can be added to the input. Physical coordinates $x$ and $y$ are normalized to $[0,1]$ and for frequencies $f = 2, 4, \ldots, 2^k$ are encoded into $4k$ channels \cite{tancik2020fourierfeaturesletnetworks}, \cite{mildenhall2020nerfrepresentingscenesneural}. For each frequency $f$, spatial information is encoded into $4$ channels with values: $\sin(2 \pi fx), \cos(2\pi f x), \sin(2 \pi fy), \cos(2 \pi f y)$, see Figure \ref{fig:coordinateEnc}.

\paragraph{Transformer.} Transformers \cite{NIPS20173f5ee243} are well suited for capturing long-term dependencies, which cannot be captured with the convolutional layers alone. We use a vision transformer encoder \cite{dosovitskiy2021imageworth16x16words}  with positional encoding. Each latent representation is divided into patches of $patch_{size} \times patch_{size}$ using a convolution with stride $patch_{size}$ and $emb_{dim}$ number of out channels, where $emb_{dim}$ is the dimensionality of the transformer token.  This embedding is then enriched by adding positional encoding that encodes the order of the patches. Since the sequence length is fixed, the positional encoding is implemented as a set of learnable parameters, each with a distinct learned value for each token position. Parameters $\lambda$ can be transformed with an MLP and used as an additional token. This is then fed as input to the transformer layers.

\paragraph{Parametric transformer.} Similarly to affine-parametrized convolutions, FiLM can be introduced after each \textit{Layer Normalization} \cite{ba2016layernormalization} in the transformer layer and on the computed query, key and value representations. To make the notation explicit, let $X_t \in \mathbb{R}^{N \times d}$ denote the token matrix obtained from the encoder output at time $t$ after patch embedding, positional encoding and, when used, parameter-token augmentation. For each attention head, let $W_Q, W_K, W_V$ denote the learned projection matrices. The corresponding query, key and value matrices are then computed as
\begin{align}
    Q_t = X_t W_Q,
    \qquad
    K_t = X_t W_K,
    \qquad
    V_t = X_t W_V.
    \label{eq:qkv_def}
\end{align}
Thus $W_Q, W_K, W_V$ are trainable parameters, while $Q_t, K_t, V_t$ are recomputed at each forward pass from the current encoded state $X_t$.

A small multi-layer perceptron maps the parameters into per-head, per-channel scaling and shifting coefficients. For example, FiLM modulation of the value representation is written as
\begin{align*}
    V_t \leftarrow V_t \odot \bigl(1 + \eta \alpha_V(\lambda)\bigr) + \eta \beta_V(\lambda),
\end{align*}
with analogous transformations for $Q_t$ and $K_t$. Here $\eta$ is a learnable per-layer parameter bounded by a fixed cap in order to prevent destabilization. When $\eta$ is close to zero, the layer behaves like standard attention, so the parameter dependence is introduced as a controlled perturbation rather than as a structural replacement.

Modulation is applied to all transformer heads, enabling the attention operator to adapt globally to the parameter vector. This allows the model to represent a parameterized family of solution operators within a unified architecture. To stabilize layer normalization under conditioning, the affine modulation networks are initialized around zero \cite{peebles2023scalablediffusionmodelstransformers}, and the modulation is written as
\begin{align*}
    x \leftarrow \mathrm{LayerNorm}(x) \odot \bigl(1 + \theta \alpha_{LN}(\lambda)\bigr) + \theta \beta_{LN}(\lambda).
\end{align*}
We set $\theta = 1\mathrm{e}{-3}$. The bounded modulation strength ensures that, at initialization, parameter-dependent modulation has negligible influence, allowing training to begin in a regime close to the unconditioned transformer. As training progresses, the network learns to scale the modulation strength appropriately. Identity initialization is used to preserve the original model behavior and to ensure stable optimization \cite{hu2021loralowrankadaptationlarge}, \cite{houlsby2019parameterefficienttransferlearningnlp}.

\begin{figure}
    \centering
\includegraphics[scale = 0.38]{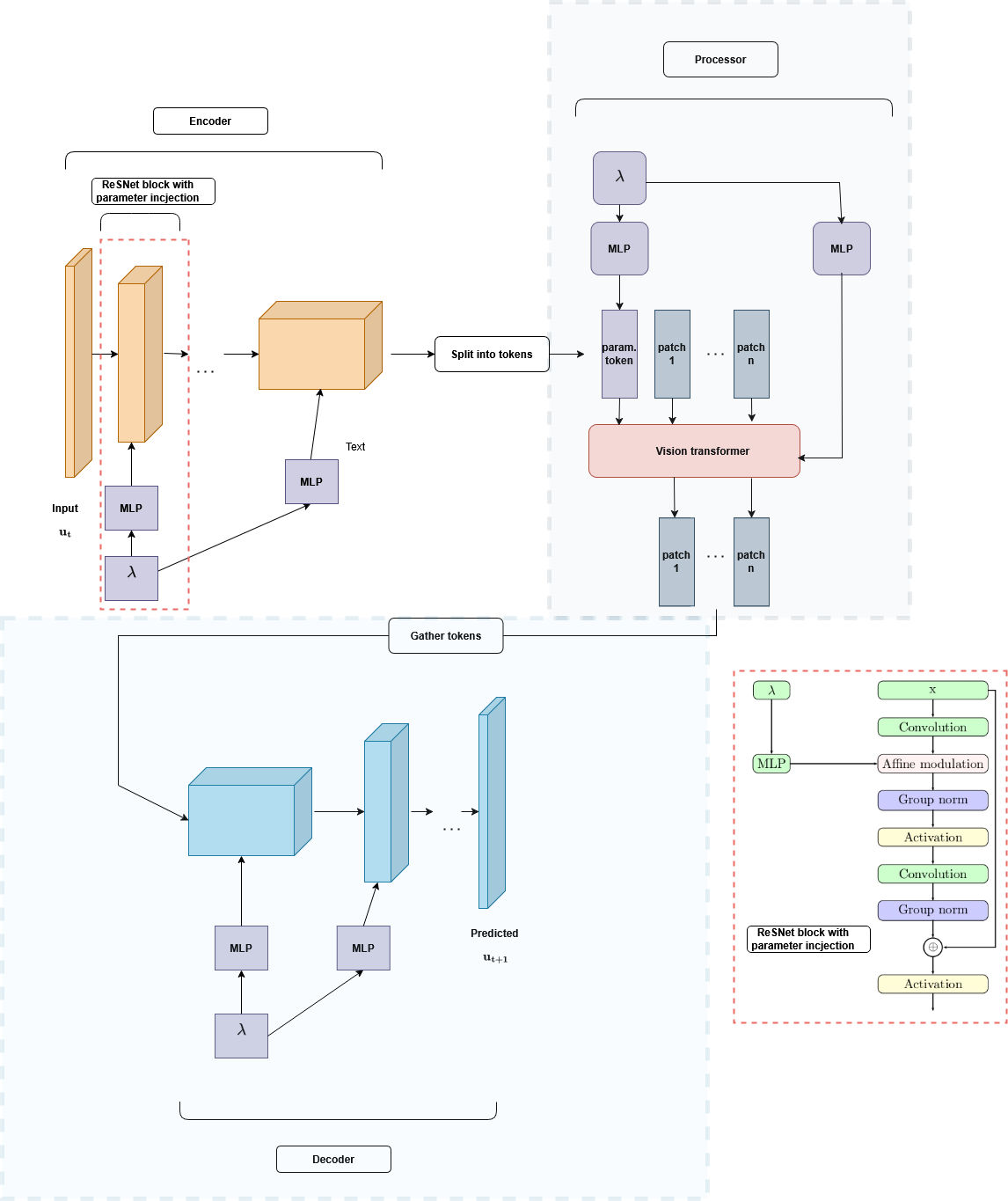}
    \caption{Basic overview of the model. The input is solution snapshot at time $\phi^t$, which is passed through convolutional encoder, then split by patches, fed through transformer encoder, whose output is then concatenated into tensor and decoded via convolutional decoder to output predicted $\phi_{t + \Delta t}$. Parameters can be injected via multi-layer perceptron into encoder, transformer and decoder via affine modulations. }
    \label{fig:model_simplified}
\end{figure}

\paragraph{Decoder.} The decoder mirrors the encoder, with ResNet blocks and optional parametric FiLMs, except for the last layers, where it outputs the solutions at time $t+\Delta t$ (without coordinate encoding) in the original resolution. 
The joint model is shown in Figure \ref{fig:model_simplified}.

The model is trained to predict solutions at time $t +  \Delta t$ from the solution at time $t$ by minimizing mean squared error. In order to improve long-term temporal prediction, scheduled sampling is used, where during training, the model observes its own predictions. Throughout this work, we used scheduled sampling with prediction $4$ steps in advance.

Concretely, starting from the ground-truth state at time $t$, the model is unrolled for $4$ consecutive steps, where at each step the input is chosen between the ground truth and the model prediction according to the scheduled sampling probability. The training loss is then defined as the average mean squared error over the entire rollout window:
\begin{align*}
    \mathcal{L} = \frac{1}{4} \sum_{k=1}^{4} \left\| \phi(t + k\Delta t) - \hat{\phi}(t + k\Delta t) \right\|^2
\end{align*}
One can also add previous time steps as preceding tokens to the input of ViT. However, in this work, we do not use any preceding tokens. For the whole simulation, prediction is done autoregressively. In all our parameter injecting modules, we do not inject time, since time will go out of training range in the time extrapolation regime. 

Despite being optimized with only short-horizon supervision (scheduled sampling length 4), the model produces long-horizon trajectories that remain accurate over hundreds of steps. This indicates that the architecture learns a stable approximation of the underlying solution operator rather than merely minimizing one-step prediction error. 

We emphasize that the proposed model could naturally be trained on inputs of varying spatial resolutions. The only required modification concerns the transformer positional encodings, which must be adapted (e.g., by using relative positional biases) when the number of tokens changes.

Since self-attention has quadratic complexity with respect to the number of tokens, the practical resolution limit is primarily determined by the token count rather than the input resolution itself. This limitation might be alleviated by employing sparse or linear-complexity attention mechanisms; however, exploring such variants is beyond the scope of this work.  

A useful way to view the transformer block is as a learned nonlocal interaction operator acting on the latent token representation. The operator-learning perspective developed in \cite{cao2021choosetransformerfouriergalerkin} is helpful in motivating this viewpoint; however, that work studies \emph{softmax-free} Fourier- and Galerkin-type attentions, whereas the present model uses standard softmax attention. For this reason, we do not claim that the formulas below follow directly from \cite{cao2021choosetransformerfouriergalerkin}. Instead, we use that reference only as motivation for interpreting attention-based architectures as nonlocal operators.
Let $z_i^t \in \mathbb{R}^d$ denote the $i$-th token in $X_t$, and let $q_i^t$, $k_i^t$ and $v_i^t$ denote the corresponding rows of $Q_t$, $K_t$ and $V_t$. Standard self-attention computes
\begin{align}
    a_{ij}^t(\lambda)
    =
    \frac{\exp\!\left( \langle q_i^t, k_j^t \rangle / \sqrt{d_k} \right)}{\sum\limits_{\ell=1}^{N} \exp\!\left( \langle q_i^t, k_\ell^t \rangle / \sqrt{d_k} \right)},
    \qquad
    {z}_i^{t+\Delta t}
    =
    \sum_{j=1}^{N} a_{ij}^t(\lambda) \, v_j^t.
    \label{eq:softmax_attention_interpretation}
\end{align}
Because $Q_t$, $K_t$ and $V_t$ are computed from $X_t$, the coefficients $a_{ij}^t(\lambda)$ depend on the current encoded state, and therefore define a state-dependent family of interaction weights on the latent grid.

\paragraph{Model interpretation.}
For linear evolution problems, one often expects a state-independent kernel associated with a Green's function or semigroup. For nonlinear problems, such as the Navier-Stokes equations considered below, the one-step solution operator is itself nonlinear, so it is natural that an effective interaction kernel in a surrogate model depends on the current state. Accordingly, we interpret the coefficients $a_{ij}^t(\lambda)$ as a learned \emph{state-dependent effective kernel} on latent tokens, rather than as a classical Green's function of the underlying PDE.

Under this interpretation, different parameter-injection mechanisms correspond to different ways of introducing parameter dependence into the learned operator. Depending on the architecture, one may distinguish the following three regimes:
\begin{enumerate}
    \item \textbf{Feature conditioning only.} The encoder and decoder depend on $\lambda$, while the transformer block itself is not explicitly conditioned on $\lambda$. In this case the latent features are parameter-dependent, but the attention rule is shared across parameters:
    \[
        u_{t+\Delta t} \approx D_{\lambda}\bigl(\mathcal{T}(E_{\lambda}(u_t))\bigr).
    \]

    \item \textbf{Attention conditioning only.} The encoder and decoder are parameter-independent, while the transformer block is conditioned on $\lambda$ through FiLM modulation, parameter tokens, or both. In this case the attention rule itself depends on the parameters:
    \[
        u_{t+\Delta t} \approx D\bigl(\mathcal{T}_{\lambda}(E(u_t))\bigr).
    \]

    \item \textbf{Fully conditioned architecture.} Both the feature extraction/reconstruction maps and the transformer block depend on $\lambda$:
    \[
        u_{t+\Delta t} \approx D_{\lambda}\bigl(\mathcal{T}_{\lambda}(E_{\lambda}(u_t))\bigr).
    \]
\end{enumerate}
The architecture used in this work is closest to the third regime, since parameter injections are allowed in the encoder, the transformer and the decoder.

The ablation results in Table \ref{adv_single_contribution} support this design choice: FiLM in the encoder and decoder (feature conditioning) and FiLM in the transformer attention (attention conditioning) achieve nearly identical error reductions individually ($0.00521$ vs.\ $0.00519$ at $T=0.4$), suggesting that neither mechanism alone is sufficient and that the fully conditioned regime benefits from both simultaneously.

 \section{Examples}
 In this section, we validate our approach on two examples. The first is the Advection-Diffusion-Reaction equation, from \cite{farenga2024latentdynamicslearningnonlinear}, which will be used as a point of comparison with previous work. The second example is the Navier-Stokes equation with the cylinder wake, which is much more challenging due to the nonlinearity of the system and different scales in the solution components. For the Navier-Stokes equations, the models will be trained on both velocity fields and pressure jointly.  
 To stabilize ViT and AE-ViT training, we use global gradient-norm clipping \cite{pmlr-v28-pascanu13} with threshold 
 $c$, meaning gradients are rescaled to have norm $c$ whenever their global norm exceeds this value. We set $c=1$ for all experiments. 
  In all trained models the learning rate follows a three-phase schedule. It increases linearly during the first $10\%$ of training steps (warm-up phase), remains constant for the subsequent $20\%$, and then decays according to a cosine annealing schedule for the remainder of training. 

 AE-ViT is compared to ViT, DL-ROM, and AE + 1D transformer. ViT receives the solution at time $t$ and PDE parameters as an additional attention token and performs one-step prediction to $t+\Delta t$. It uses the same number of attention layers and patch size as AE-ViT, but unlike AE-ViT, it does not reduce the input resolution, resulting in more tokens, and it does not employ a convolutional encoder/decoder or coordinate encoding. DL-ROM combines a vanilla autoencoder to obtain a latent representation of the solution and a fully-connected neural network to map from PDE parameters to the latent space, predicting the entire solution trajectory simultaneously. Following standard practice \cite{Franco_2022}, parameters are not injected into the autoencoder and no coordinate encoding is used. AE + 1D transformer consists of a vanilla autoencoder and a 1D transformer applied to the latent representation to model temporal evolution, performing one-step prediction like ViT, without parameter injection in the autoencoder. In contrast, AE-ViT reduces the input resolution before the ViT module, resulting in fewer tokens, and uses a fully convolutional encoder and decoder, reducing the number of trainable parameters compared to the baselines. AE-ViT also incorporates PDE parameters and employs coordinate encoding, which enables efficient modeling of temporal evolution.

We employ component-wise z-score normalization (zero mean, unit variance), with statistics computed per channel over the training dataset. Model parameters are normalized separately using their own dataset-level statistics. All reported errors are computed after transforming predictions back to the original (unnormalized) scale.

 Models are compared according to relative error, where one step relative error between $\phi^t$ and predicted $\hat{\phi}^t$ is defined as 
 \begin{align*}
     l(\hat{\phi}^t, \phi^t) = \frac{\sqrt{ \sum\limits_{i,j}(\hat{\phi}^t - \phi^t)_{ij}^2}}{\sqrt{ \sum\limits_{i,j}( \phi^t)_{ij}^2}},
 \end{align*}
 where $(\phi^t)_{ij}$ and $(\hat \phi^t)_{ij}$ are the exact solution and prediction of the network on position $(i,j)$ at time $t$, respectively. Note that this is a discrete analogue of $\frac{\|\phi(t.,)-\hat{\phi}(t,.)\|_{L^2(\Omega(\lambda))}}{\|\phi(t,.)\|_{L^2(\Omega(\lambda))}}$.
 For evaluation of all models, we use the relative rollout error:
 \begin{align}
     \frac{1}{N_t} \sum_{k=1}^{N_t} l(\hat{\phi}^{k\Delta t}, \phi^{k\Delta t}),
     \label{relative}
 \end{align}
 where $\hat{\phi}_{k\Delta t}$ is prediction of the model at time $k \Delta t$, calculated autoregressively from initial condition and parameters, and $N_t$ is the number of time steps. This is a discrete analogue of $L^1(0,N_t\Delta t;L^2(\Omega(\lambda)))$ norm.
 In case of the system of equations, the relative error is evaluated for each solution component separately.

 \subsection{Advection-Diffusion-Reaction}
 For our first example, \cite{farenga2024latentdynamicslearningnonlinear} is closely followed. 
 Namely, the equation is
 \begin{align}
\begin{cases}
    \phi_t - \nabla \cdot (\mu_1 \nabla \phi) + b(t) \cdot \nabla \phi + \phi  = f(\cdot; \mu_2,  \mu_3),  & {\rm in\;\;}\Omega \times (0,T]   \\
     \mu_1 \nabla \phi \cdot n  = 0 & {\rm in\;\;}\partial \Omega \times (0,T],  \\
      \phi(x,0) = 0  & {\rm in\;\;} \Omega, \\
\end{cases}
 \label{sub_adv}
\end{align}
where $b(t)= [\cos(t), \sin(t)]$, $f(\textbf{x};\mu_2,\mu_3)=10\exp(-((x-\mu_2)^2+(y-\mu_3)^2)/0.07^ 2)$. $\Omega$ is unit square. All simulations are calculated up to $T = 10\pi$ using $1000$ time steps. The equation has been solved with FEM using $\mathbf{P1}$ elements, using $32 \times 32$ grid, resulting in $1024$ degrees of freedom (DoFs). 

The sampled parameters  $(\mu_1, \mu_2, \mu_3) \in [0.02, 0.05] \times [0.4, 0.6]^2$ are partitioned into $\mathcal{P}_{train}$,$\mathcal{P}_{valid}$, and $\mathcal{P}_{test}$. The training set consists of $800$ simulations up to $T_{train} = 4\pi$, or $400$ time steps,  the  validation consists of $200$ simulations, each up to $T= 4\pi$,  test set consists of $200$ simulations, all simulated until time $10 \pi$. 

We fix the architecture for AE-ViT and vary the learning rate and weight decay, as shown in Table \ref{ae_vit_adv_abl}.
\begin{table}[]
    \centering
    \begin{tabular}{|c|c|}
            \hline
        Hyperparameter name & Values \\
        \hline
        Kernels per layer & [$32, 64, 64, 128, 256$ ]  \\
                \hline

        Strides per layer & [ $1, 2, 1, 1, 1$] \\
                \hline

        Nbr of transformer layers & $4$ \\
                \hline

        Learning rate &  $1\text{e-}5, 3\text{e-}5, 1\text{e-}4, 3\text{e-}4, 1\text{e-}3$ \\
                \hline

        Weight decay & $1\text{e-}4, 1\text{e-}3, 1\text{e-}2$ \\
        \hline
        Embedding dimension & $256$ \\
        \hline
        Patch size & $2 \times 2$ \\
        \hline
        Encoder latent size & $16 \times 16 \times 256$ \\
        \hline
    \end{tabular}
    \caption{AE-ViT hyperparameter search. We fix the architecture and vary learning rate and weight decay. The autoencoder has  5 hidden layers with $32, 64, 64, 128, 256$ kernels respectively and strides $1, 2, 1, 1, 1$. The decoder mirrors the encoder. The transformer has $4$ layers, with embedding dimension $256$ and splits encoded snapshot into $2 \times 2$ patches, resulting in $64$ patches per snapshot. The encoder produces a latent representation of size $16 \times 16 \times 256$. We fix the architecture and find the best combination of learning rates and weight decays, where tried learning rates are $1\text{e-}5, 3\text{e-}5, 1\text{e-}4, 3\text{e-}4, 1\text{e-}3$ and weight decays $1\text{e-}4, 1\text{e-}3, 1\text{e-}2$ with $2$ different (but fixed) random seeds, which determine network initialization, training data ordering and training. Performance of each configuration is consistent across seeds. }
    \label{ae_vit_adv_abl}
\end{table}
  Our attempt to train the autoencoder and ViT separately was unsuccessful. More precisely, after the autoencoder was trained, ViT training on the autoencoder's fixed latent representations would not converge. We assume it is because of high-dimensional latent representation and possible mismatch between spatial encoding and latent evolution.  
  
In order to fairly compare our additions to the architecture, we start with the baseline model, which consists of the encoder, ViT and the decoder, with parameters injected only as a ViT context token. After finding the best combination of learning rate and weight decay, we performed an ablation study on AE-ViT with different combinations of types of parameter injection and coordinate encoding to check the impact of proposed changes in our architecture and investigate its individual and combined contribution to the model performance. Ablation study specifications are given in Table \ref{ablation_config}.
 \begin{table}[]
     \centering
     \begin{tabular}{|c|c|}
        \hline
         coordinate encodings, nbr of Fourier frequencies & 0, 4  \\
         \hline
         parameter injection in the encoder and decoder & True, False \\
         \hline
         parameter injection in the transformer layer normalization & True, False \\
         \hline
         parametric token & True, False \\
         \hline
         parametric FiLM of query, key and value matrices & True, False\\
         \hline
     \end{tabular}
     \caption{Hyperparameters varied in the ablation study. We investigate effect of parameter injection through multiple places in the architecture and effect of having positional encoding.}
     \label{ablation_config}
 \end{table} 
 
 For training the DL-ROMs and the latent transformer we use the same convolutional structure in the encoder and the decoder as in our method. For DL-ROMs and latent transformers, encoder and decoder have additional linear layers for projecting latent tensor to latent vector and vice versa. In the latent transformer, we experiment with adding preceding tokens (of 32 snapshots), or without preceding tokens, as in our model. Note that most of the parameters for these models belong to the projection of the latent tensor to a latent vector. 
 
For DL-ROM and latent transformer, we choose the autoencoder as the best among models hyperparameter by doing a grid-search, see Table \ref{adv_ae_abl}.
\begin{table}[]
    \centering
    \begin{tabular}{|c|c|}
    \hline 
     Hyperparameters & Values \\ 
     \hline
      Kernels per layer   & [$32, 64, 64, 128, 256$], [$64, 128, 128, 256, 512, 512$]  \\
          \hline
      Stride size per layer   & [$1, 2, 1, 1, 1$], [$1, 2, 2, 2, 2, 1$]  \\
          \hline
      Latent dimension & $32, 64, 128, 256$ \\
        \hline
      Learning rate & $3\text{e-}4, 1\text{e-}4$ \\
        \hline
      Weight decay & $1\text{e-}3, 1\text{e-}2$ \\
      \hline
      
    \end{tabular}
    \caption{Autoencoder hyperparameter grid-search specifications. We trained multiple architectures, either with 6 layers with $64, 128, 128, 256, 512, 512$ kernels and strides $1, 2, 2, 2, 2, 1$ or with $5$ layers with $32, 64, 64, 128, 256$ kernels respectively and strides $1, 2, 1, 1, 1$, with varying latent dimensions $32, 64, 128, 256$ and learning rates $3\text{e-}4, 1\text{e-}4$, weight decays $1\text{e-}3, 1\text{e-}2$ and $5$ different seeds for each configuration, resulting in $160$ models.}
    \label{adv_ae_abl}
\end{table}
We observed that the best model in terms of the reconstruction error has latent dimension $256$ and number of kernels and strides as in our AE-ViT model. This autoencoder is used for spatial compression for DL-ROM and 1D transformer. For the fully-connected network in DL-ROM that maps parameters to its latent representations, we used $6$ hidden layers. We trained a series of models with $512$, $1024$, $2048$, and $4096$ neurons per layer and stopped at $4096$ since validation relative error stopped decreasing. The reported relative error is for the model having $4096$ neurons per layer.
 
For the 1D transformer, we choose the transformer hyperparameters as the best among the models hyperparameters by doing a grid-search, see Table \ref{ablation_adv_transformer}. The best model has $8$ layers and no preceding tokens,  feedforward transformer dimension of $1024$, scheduled sampling window of length $8$, learning rate $3\text{e-}4$ and weight decay $1\text{e-}2$.

For ViT, the same transformer configuration as in AE-ViT is used.

\begin{figure}
    \centering
    \includegraphics[width=0.5\linewidth]{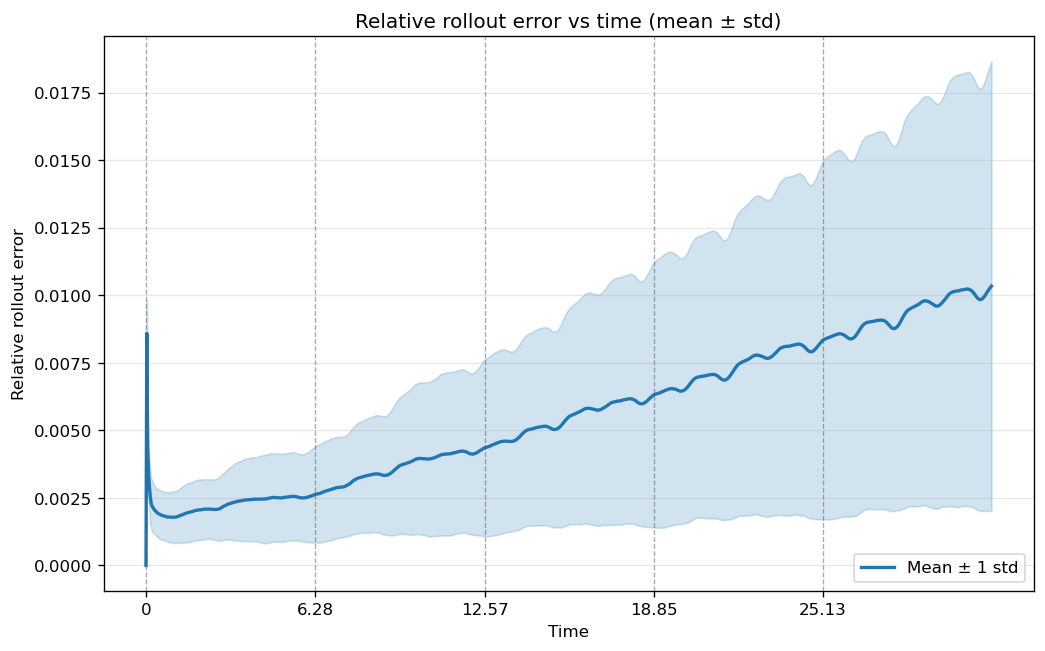}
    \caption{Mean relative rollout errors per step on test set (solid line), with standard deviation (lighter area). Relative rollout error grows linearly. Standard deviation increases as time progresses. }
    \label{adv_roll}
\end{figure}

\begin{figure}
    \centering
    \includegraphics[width=0.3\linewidth]{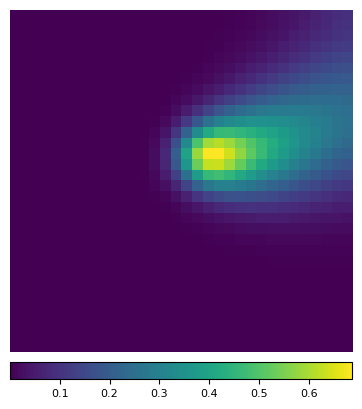}
    \includegraphics[width=0.3\linewidth]{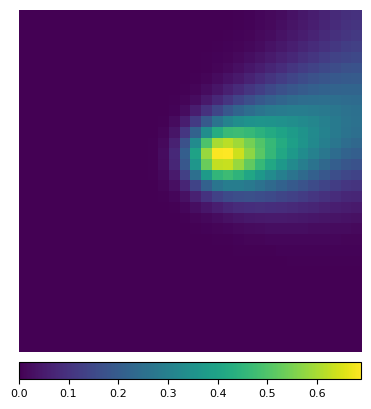}
    \includegraphics[width=0.3\linewidth]{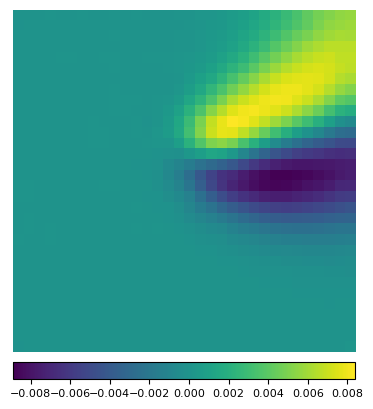}
    \includegraphics[width=0.3\linewidth]{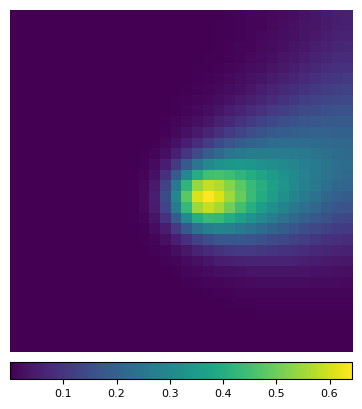}
    \includegraphics[width=0.3\linewidth]{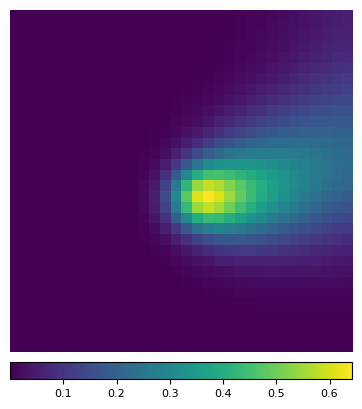}
    \includegraphics[width=0.3\linewidth]{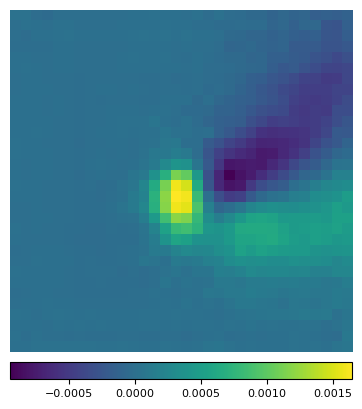}
    \caption{Rollout results after 1000 steps. First row: correct solution (left), prediction (middle), pointwise error(right) for parameters $\mu_1 = 0.0206, \mu_2 = 0.544, \mu_3 = 0.428$. Second row: correct solution (left), prediction (middle), pointwise error (right) for parameters $\mu_1 = 0.0264, \mu_2 = 0.528, \mu_3 = 0.553$. }
    \label{adv_pred}
\end{figure}

\begin{table}[]
    \centering
    \begin{tabular}{|c|c|}
    \hline
      Number of transformer layers & 4, 8 \\
      \hline
      Transformer embedding dimension   & 256 \\
      \hline
      Feedforward transformer dimension   & 512, 1024\\
      \hline
      Scheduled sampling window & 4, 8 \\
      \hline
      Preceding tokens length & 0, 32\\ 
      \hline
      Learning rate & $1\text{e-}3$, $3\text{e-}4$ \\
      \hline
      Weight decay &  $1\text{e-}2$, $1\text{e-}3$ \\
      \hline
    \end{tabular}
    \caption{Latent transformer hyperparameter grid search specifications. Learning rate was $1\text{e-}3$, $3\text{e-}4$, weight decay $1\text{e-}2$, $1\text{e-}3$, transformer embedding dim $256$, feedforward transformer dimension $512$ or $1024$, with $4$ or $8$ transformer layers, with scheduling sampling window $4$ or $8$. Models were trained either without preceding tokens, or preceding tokens of $32$ prior latent codes. The best model has $8$ layers, no preceding tokens, feedforward transformer dimension of $1024$, scheduled sampling window of length $8$, learning rate $3\text{e-}4$ and weight decay $1\text{e-}2$. Each configuration is trained across $4$ different random seeds, resulting in $128$ models }
    \label{ablation_adv_transformer}
\end{table}
Comparison of models is given in Table \ref{adv_comparison}. We observe that our AE-ViT significantly outperforms other models. 
\begin{table}[]
    \centering
    \begin{tabular}{ |c|c|c|c|}
    \hline
        model & relative rollout error (T=0.4) & relative rollout error (T=1.0) & parameter count\\
        \hline
        AE-ViT (ours) & \boldsymbol{$0.0029$} & \boldsymbol{$0.0059$} & \boldsymbol{$\approx 6$} million\\ 
        \hline
        ViT & $0.0997$ & $0.2366$ &  $\approx 4$ million\\
        \hline
        DL-ROM & $0.0123$ & $0.6473$ & $\approx 52$ million\\
        \hline
        AE + 1D transformer & $0.0117$ & $0.0229$& $\approx 38$ million\\
        \hline
    \end{tabular}
    \caption{Comparison of different models in the Advection-Diffusion-Reaction case. We report only the best model performance for each instance. Across runs with different random seeds, the average of the best validation errors aligns with the overall best-performing model. Not only does AE-ViT outperform other autoencoder-based models, but it also has way fewer parameters. Furthermore, we see that the relative rollout error after the training window has the slowest increase in AE-ViT. Mean rollout error is reported for time intervals $[0,0.4]$ ($2$ periods) and $[0,1]$ ($5$ periods). Most parameters in AE+1D transformer and DL-ROM are in fully connected layers that map from latent 2D to a latent vector and vice versa. DL-ROM is not able to extrapolate in time since the time goes out of the training distribution. Transformer-based models have a similar rate of mean relative rollout error increase between different reported time intervals.}
    \label{adv_comparison}
\end{table}
To show the model's long-term stability, we plot the relative rollout error over time steps. Results are in Figure \ref{adv_roll}, where mean relative rollout errors over time steps and standard deviation are shown. The relative rollout error increases approximately linearly, while the recurring dips in the rollout error are consistent with the periodic behavior of the solution after the transient. In Figure \ref{adv_pred} we show the prediction for a specific sample. 

We analyze contributions of our changes in the ablation study. The varied hyperparameters are FiLM in the attention, FiLM in the encoder and decoder, FiLM in the transformer layer normalization and parameters as an additional token. The number of frequencies in input coordinate encoding is either $0$ or $4$, resulting in $32$ different combinations. All combinations are run with $5$ fixed random seeds, resulting in $160$ trained models. Since in the Advection-Diffusion-Reaction example, all trajectories start from the same initial condition, the error of the baseline AE-ViT without any parameter injection will naturally be big since the model cannot differentiate between the trajectories. For this study, we report the effect of adding parameters in comparison with baseline AE-ViT with rollout starting from the first step. In order to demonstrate that each of our proposed changes greatly reduces rollout relative error, we also report errors for models with only one enhancement, see Table \ref{adv_single_contribution}.

\begin{table}[]
    \centering
    \begin{tabular}{|c|c|c|}
        \hline
        Model & mean relative rollout error $$(T=0.4)$$ & mean relative rollout error $$(T=1.0)$$  \\
        \hline
        Baseline &     $0.4274$                &    $0.6330$                    \\
        \hline
        Coordinate encodings &  $0.0575$                   &   $0.1317$                     \\
        \hline
        FiLM in encoder and decoder &       $0.005207 $             &    $0.010747$	                   \\
        \hline
        FiLM in transformer layer normalization &   $0.15602$                  &       $0.31867$                 \\
        \hline
        FiLM in transformer attention &  \boldsymbol{$0.005188$}                   &    \boldsymbol{$0.009559$}                   \\
        \hline
        Parameters token & $0.008613$ & $0.016004$ \\
        \hline
        \end{tabular}
    \caption{Mean relative rollout error comparison of baseline AE-ViT with respect to only one enhancement. We see that each of the proposed parameter injections and coordinate encodings greatly reduce the relative error, with the largest mean relative rollout error reduction with FiLM in transformer attention. }
    \label{adv_single_contribution}
\end{table}

\subsection{Navier-Stokes flow around an obstacle}
Further, we examine our model on the 2D Navier-Stokes equations in $\Omega(\lambda)$, where  $\Omega(\lambda)$ is a rectangular pipe of length $5$ and width $1$ with a circular obstacle. 
\begin{align}
    \begin{cases}
        \mathbf u_t + \mathbf  u \cdot \nabla \mathbf  u  + \nabla p = \frac{1}{Re}\Delta \mathbf  u  \text{ in } \Omega(\lambda)\\
        \nabla \cdot \mathbf  u = 0  \text{ in } \Omega(\lambda) \\
        \mathbf  u(0, x, y) = 0 \\
        \mathbf u = 0  \text{ on $\Gamma_{bot}, \Gamma_{top}$ and the edge of the circle with center $(x_c, y_c)$ and radius $r$} \\
        \sigma \textbf{n} = 0 \text{ for x = 5} \\
        \mathbf  u(t,0,y) = 4(1 + A\sin(2\pi f t))y(1-y),
    \end{cases}
    \label{sub_ns}
\end{align}
where $\Gamma_{top}, \Gamma_{bot}$  are the top and bottom sides of the rectangle, $\sigma$ is the fluid stress tensor and \textbf{n} is the unit normal. The parameters are: magnitude of time-periodic perturbation $A \in [0.05, 0.30]$, center of the circle $(x_c, y_c) \in [0.9,1.3] \times [0.4, 0.6]$, circle radius $r \in [0.06, 0.12]$, inflow frequency $f=0.74$ is fixed for all simulations. 

The Reynolds number $Re$ is defined as the ratio of inertial to viscous forces in the flow and is given by $Re = \frac{UL}{\nu}$, where $U$ is the characteristic velocity, $L$ is the characteristic length scale, and $\nu$ is the kinematic viscosity. In flow past a circular obstacle, the natural length scale is the obstacle size, here taken as radius $r$. In this study, the Reynolds number depends on the circle radius as $Re \in [\frac{180}{4r}, \frac{540}{4r}]$. This range was selected so that all considered cases exhibit vortex shedding and nontrivial wake dynamics.

All solutions are interpolated to a $64 \times 320$ rectangular grid and masked by the domain characteristic function in order to use the convolutional encoder and decoder. Since both velocities have $0$ as a boundary condition on the obstacle edge, multiplying by the mask does not produce discontinuities in velocity fields, but due to the nature of the problems, gradients near the obstacle are sharp. Simulations are calculated with time step $dt=0.002$, where results are saved every third step. Simulations are run for $10$ inlet periods to allow the initial transient to decay, after which $5$ periods are saved. We use $2$ periods for training, resulting in $450$ snapshots per training simulation. $800$ different simulations are used for the training set. Our model learns the velocities in the $x$ and $y$ directions and pressure jointly. We train the model for $2$ periods (up to $T=2.7$). Additionally, we evaluate the rollout starting from $T=0$ over a total of $5$ periods, up to $T=6.75$.

Due to the larger computational cost compared to the Advection-Diffusion-Reaction example (see equation (\ref{sub_adv})), we further restrict our hyperparameter search to AE-ViT with all proposed parameter injections and $4$ coordinate encoding frequencies used in every search instance. First, we sweep over $5$ different random seeds, learning rates $1\text{e-}4, 3\text{e-}4, 1\text{e-}3$, weight decays $1\text{e-}4, 1\text{e-}3, 1\text{e-}2$, with an encoder with $5$ layers with $32,64,64,128,256$ kernels per convolutional layer and respective strides $1, 2, 2, 2, 1$, and ViT patch size $2$. Models with learning rate $3\text{e-}4$ and weight decay $1\text{e-}4$ have the smallest relative rollout error on the validation set. We report on the relative test error for all channels in the training window $T=2.7$ and for additional $3$ periods until $T=6.75$. The best results are obtained with AE-ViT (ours) in all channels. 

\begin{figure}
    \centering
    \includegraphics[width=0.3\linewidth]{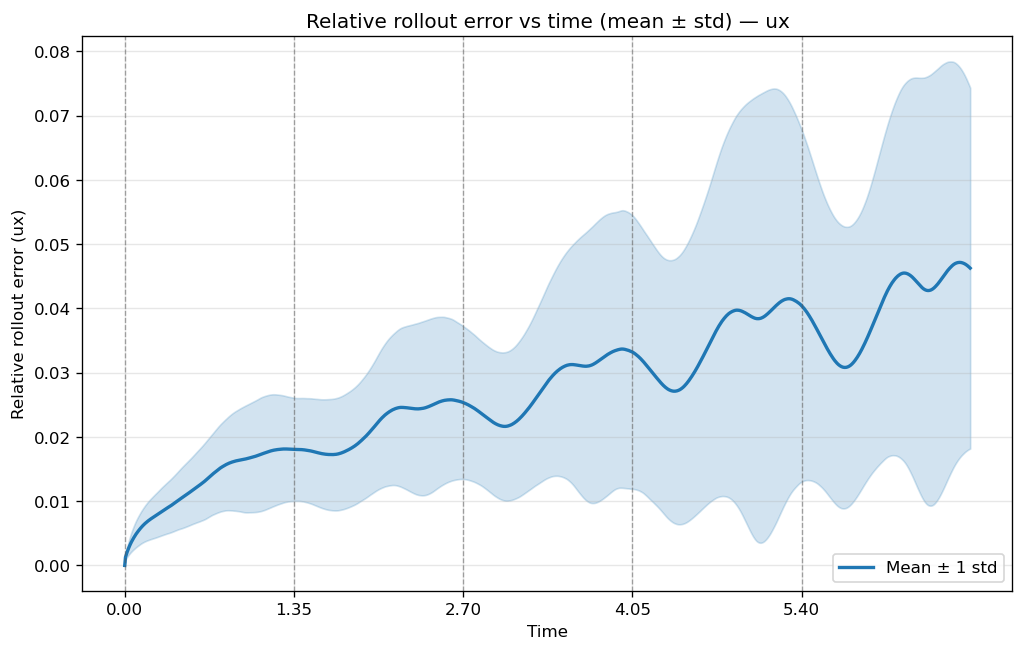}
    \includegraphics[width=0.3\linewidth]{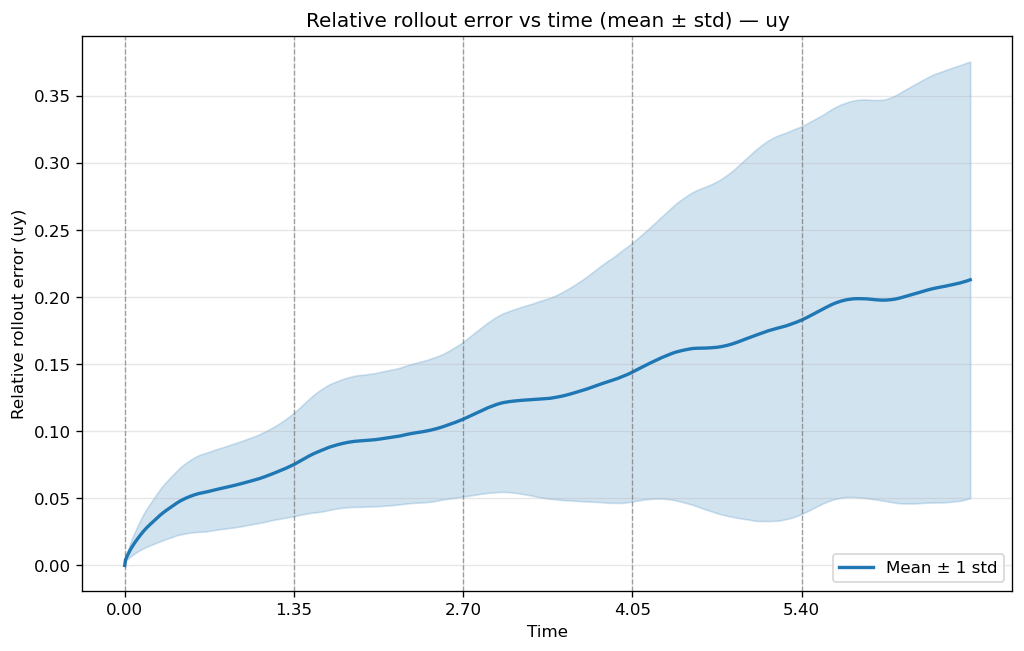}
    \includegraphics[width=0.3\linewidth]{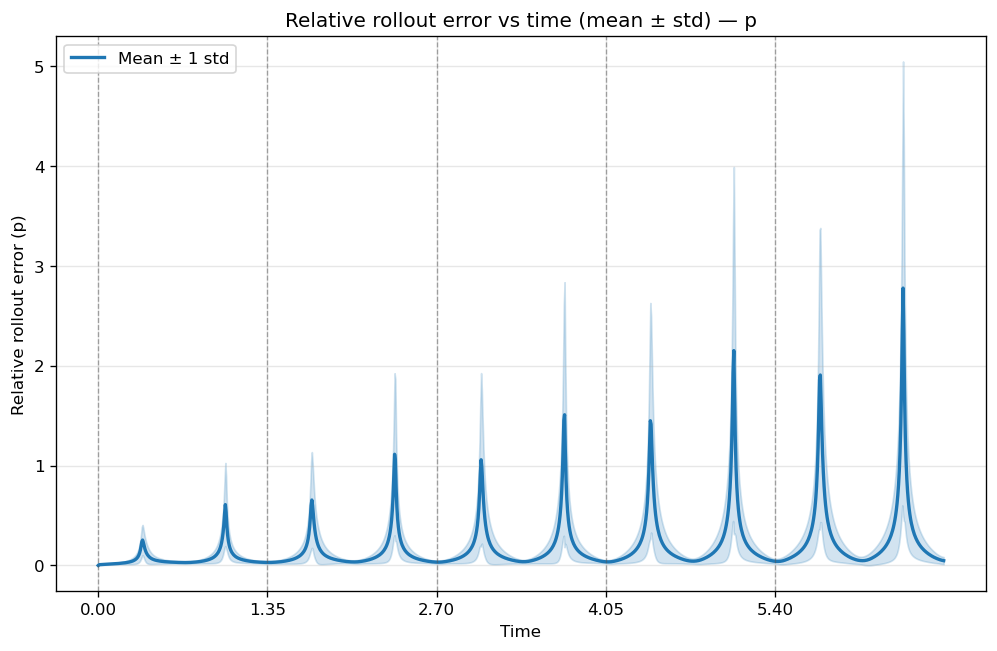}
    \caption{Mean relative rollout error (solid line) with standard deviation (shaded area) on the test set over time for velocities $u_x$ (left), $u_y$ (middle), and pressure (right). Relative rollout error grows linearly for velocities $u_x$, $u_y$ but has periodic spikes for pressure. Standard deviation increases as time progresses. Unlike velocity, pressure in incompressible flow is determined globally through a Poisson equation, making it more sensitive to phase errors in the periodic shedding cycle. A small phase drift in the predicted vortex positions leads to large pointwise pressure errors at shedding events, even when the overall flow structure is well captured. }
    \label{ns_rollout}
\end{figure}

\begin{figure}
    \centering
    \includegraphics[width=0.3\linewidth]{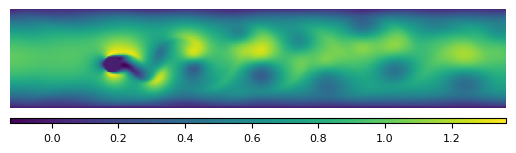}
        \includegraphics[width=0.3\linewidth]{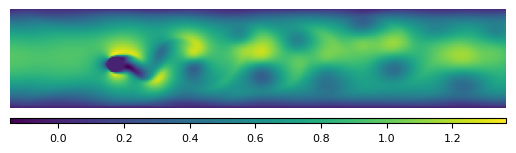}
    \includegraphics[width=0.3\linewidth]{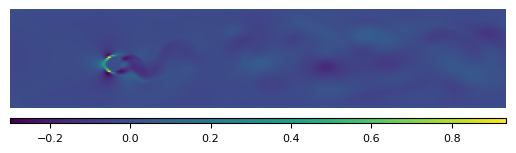}
    \includegraphics[width=0.3\linewidth]{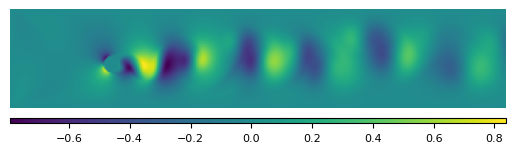}
        \includegraphics[width=0.3\linewidth]{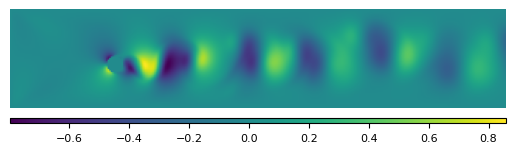}
    \includegraphics[width=0.3\linewidth]{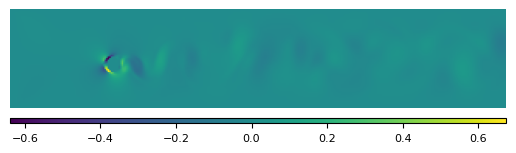}
    \includegraphics[width=0.3\linewidth]{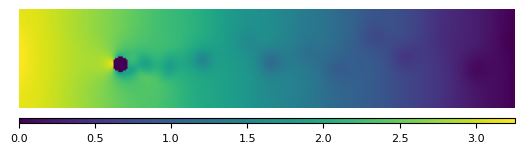}
        \includegraphics[width=0.3\linewidth]{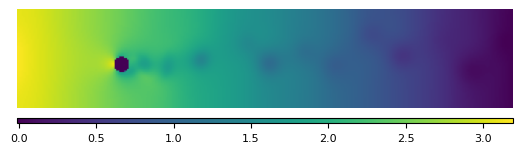}
    \includegraphics[width=0.3\linewidth]{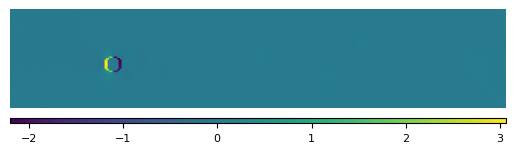}

    \caption{Prediction results. Reference solution (left column), network prediction (middle column), and pointwise error (right column) for $Re= 1.557e+03, A = 1.9489e-01, x_c =1.019, y_c= 5.637e-01,
       r =  7.772e-02$ at time $T=6.75$. Velocity in the $x-$direction is in the first row, velocity in the $y-$direction is in the second row, and pressure is in the third row.}
    \label{ns_fig}
\end{figure}

For the autoencoder-based models (DL-ROM, AE +  1D transformer), we first choose the autoencoder with the lowest mean validation relative loss (equation (\ref{relative})). Hyperparameters are shown in Table \ref{NS_AE_ablation}. The best autoencoder in terms of relative reconstruction error is the one with learning rate $3\text{e-}4$, weight decay $1\text{e-}2$ and latent dimension $256$. DL-ROM and transformer are trained on latent representations. The mean and standard deviation for each solution component and for each time step in the rollout are shown in Figure \ref{ns_rollout}. Visual results for our model are shown in Figure \ref{ns_fig}. The results of comparison of all models involved are in Table \ref{ns_comparison}.
\begin{table}[]
    \centering
    \begin{tabular}{|c|c|}
        \hline
        Kernels per layer & [$32,64,64,128,256$] \\
        \hline
        Strides per layer &  [$1, 2, 2, 2, 1$]\\
        \hline
        Learning rate & $1\text{e-}4, 3\text{e-}4$  \\
        \hline
        Weight decay & $1\text{e-}3, 1\text{e-}2$ \\
        \hline
        Latent dimension & $ 64, 128, 256$ \\ 
        \hline
    \end{tabular}
    \caption{Ablation study for the autoencoder models. All trained autoencoders have $32, 64, 64, 128, 256$ convolutional kernels per layer and respective strides of $1, 2, 2, 2, 1$. We vary the learning rate to be either $1\text{e-}4, 3\text{e-}4$, weight decay $1\text{e-}3$ or $1\text{e-}2$, latent dimension $64$, $128$ or $256$. We train all combinations with $5$ different random seeds, so in total $60$ models, which are trained jointly on all channels. The best autoencoder in terms of mean validation relative reconstruction error over seeds is the one with learning rate $3\text{e-}4$, weight decay $1\text{e-}2$ and latent dimension $256$.  }
    \label{NS_AE_ablation}
\end{table}

\begin{table}[]
    \centering
    \begin{tabular}{ |c|cc|cc|cc|}
    \hline
 & \multicolumn{2}{c|}{$u_{x}$} 
 & \multicolumn{2}{c|}{$u_{y}$} 
 & \multicolumn{2}{c|}{$p$} \\
\cline{1-2} \cline{2-3} \cline{4-5} \cline{6-7}
        model & $T=2.7$ &  $T=6.75$  & $T=2.7$ &$T=6.75$ &  $T=2.7$  & $T=6.75$  \\
        \hline
        AE-ViT (ours)  & \boldsymbol{$0.01725$}  & \boldsymbol{$0.0276$} & \boldsymbol{$0.0724$} &  \boldsymbol{$0.1271$} &\boldsymbol{$0.0999$}&  \boldsymbol{$0.1861$} \\
        \hline
        ViT & $0.3876$ & $0.4330$ & $1.0225$ & $1.0222$& $1.6385$ & $1.8081$  \\
        \hline
        DL-ROM & $0.1514$ & $0.4147$ & $1.0104$&  $1.3706$ & $0.9318$ & $3.3751$      \\
        \hline
        AE + 1D transformer & $0.0839$ & $0.1391$ & $0.4194$ & $0.7166$ & $0.9068$ & $1.3438$ \\
        \hline
    \end{tabular}
    \caption{Comparison of different models. Models are trained up to $T = 2.7$. Mean rollout relative error for each of the components of the solution of Navier-Stokes is reported for time intervals $[0, 2.7]$ and $[0, 6.75]$. For models that were trained on one-step prediction (all except DL-ROM), the reported error is pure rollout error, starting only with initial conditions and parameters.  }
    \label{ns_comparison}
\end{table}

\section{Conclusion}
In this work, we have developed a new architecture for autoregressive parametric PDE evolution, combining the strengths of autoencoders for resolution reduction and vision transformers for capturing long-range spatial interactions. We demonstrated that time evolution can be trained effectively and that the model is capable of joint convolutional autoencoder and vision transformer training, which is an advance over many existing approaches. Our proposed parameter injection and coordinate encoding greatly enhance the prediction accuracy. In the challenging example of estimating the velocity for the flow around a cylinder obstacle (see equation (\ref{sub_ns})), our proposed model achieves around $5$ times lower relative error than the best of the alternative methods. Such a model is stable in the sense that the relative error accumulates approximately linearly, even for 250 times more steps than the scheduled sampling window length, thus significantly reducing the training computational cost. The main limitations of the proposed method are dependence on the interpolation of the solutions to a rectangular domain, so it is not appropriate for domains that do not fit naturally in rectangular domains, and quadratic computational cost in the transformer layer.
The future direction of this research involves mitigating these limitations and developing error and complexity bounds of our model with the use of neural network approximation theory.    

\section{Acknowledgments}
This research was carried out using the advanced computing service provided by the University of Zagreb University Computing Centre - SRCE. This research was supported by the Croatian Science Foundation under the project number IP-2022-10-2962. BM was supported by the European Union – NextGenerationEU through the National Recovery and Resilience Plan 2021-2026. Institutional grant of University of Zagreb Faculty of Science IK IA 1.1.3. Impact4Math.

\bibliographystyle{unsrt}  
\bibliography{references}

\end{document}